\documentclass[journal]{IEEEtran}
\usepackage{amsmath,amssymb,graphicx,booktabs,array,cite,xcolor}
\usepackage{tikz}
\usetikzlibrary{arrows.meta,positioning,calc,fit,backgrounds}
\usepackage[hidelinks]{hyperref}
\newcommand{\vit}{ViT\textsuperscript{3}}
\newcommand{\model}{ChangeT\textsuperscript{3}}

\newcommand{\cat}{\operatorname{Cat}}
\newcommand{\up}{\mathcal{U}}
\newcommand{\silu}{\operatorname{SiLU}}
\newcommand{\dw}{\operatorname{DWConv}}
\newcommand{\LN}{\operatorname{LN}}
\newcommand{\CE}{\operatorname{CE}}
\newcommand{\authorblock}[2]{%
  \parbox[t]{0.32\textwidth}{\centering
    {\sublargesize #1}\\[0.3ex]
    \normalsize\textit{University of Peradeniya}\\
    Peradeniya, Sri Lanka\\
    #2}}

\newcommand{\FoneTthree}{81.64}

\newcommand{\FoneTthreeMean}{80.61}
\newcommand{\FoneTthreeStd}{0.90}

\newcommand{\FoneTthreeZero}{78.73}
\newcommand{\FoneGlobalOnly}{80.18}
\newcommand{\FonePre}{85.25}
\newcommand{\FonePreLD}{85.31}

\newcommand{\DeltaFullNone}{2.91}
\newcommand{\DeltaGlobalNone}{1.45}

\newcommand{\ParamsTthree}{40.28}
\newcommand{\ParamsCF}{41.03}
\newcommand{\FoneCFMean}{78.84}
\newcommand{\FoneCFStd}{0.21}

\newcommand{\DeltaMeanMeanCF}{1.77}
\newcommand{\FoneTthreeZeroMean}{78.63}
\newcommand{\FoneTthreeZeroStd}{0.20}

\newcommand{\DeltaMeanMeanNone}{1.97}
\newcommand{\FoneLocalOnly}{80.80}
\newcommand{\DeltaLocalNone}{2.07}
\newcommand{\FoneCFPre}{83.98}
\newcommand{\XDCF}{6.64}
\newcommand{\XDTthree}{1.83}
\newcommand{\XDTthreeZero}{3.57}
\newcommand{\XDPreLD}{13.90}
\newcommand{\XDCFPre}{9.23}
\newcommand{\NSYSU}{4{,}000}
\newcommand{\NDSIFN}{48}
\newcommand{\NWHU}{762}
\newcommand{\SYSUPrecTthree}{0.68}
\newcommand{\SYSURecTthree}{0.006}
\newcommand{\SYSUGTFrac}{23.6}
\newcommand{\SYSUPredFracTthree}{0.20}
\newcommand{\GFLOPsCF}{277.5}
\newcommand{\GFLOPsTthree}{284.4}
\newcommand{\LatCF}{16.2}
\newcommand{\LatTthree}{89.9}

\newcommand{\ThrCF}{79.1}
\newcommand{\ThrTthree}{50.4}

\newcommand{\ScaleRatioLow}{5.1}
\newcommand{\ScaleRatioHigh}{1.7}
\newcommand{\DriftLatX}{2.4}
\newcommand{\DriftThrPct}{15}
\newcommand{\InnerLatPct}{49}
\newcommand{\InnerLatPctBEight}{23}

\newcommand{\EffGPU}{RTX 6000 Ada Generation}
\newcommand{\NTrain}{10{,}192}

\newcommand{\NTest}{5{,}568}
\newcommand{\NTrainParents}{637}
\newcommand{\NTestParents}{348}
\newcommand{\PatchSize}{256}
\newcommand{\Iters}{50k}
\newcommand{\BatchSize}{8}
\newcommand{\LR}{1\times10^{-4}}
\newcommand{\WD}{0.01}
\newcommand{\EvalEvery}{1{,}274}
\newcommand{\BndTol}{2}
\newcommand{\DSWeights}{0.5, 0.5, 0.5, 0.8, 1}
\newcommand{\Seed}{42}

\begin{document}
\title{\model: Evaluating Vision Test-Time Training for Remote Sensing Change Detection}
\author{%
  \vspace*{1em}%
  \hspace*{1.5em}\authorblock{Sharukshan Niranjan}{n.sharukshan@gmail.com}%
  \authorblock{Oshadha Samarakoon}{e21345@eng.pdn.ac.lk}%
  \authorblock{Nichula Wasalathilaka}{e20425@eng.pdn.ac.lk}\\[2ex]%
  \authorblock{Roshan Godaliyadda}{roshang@eng.pdn.ac.lk}%
  \authorblock{Parakrama Ekanayake}{mpbe@eng.pdn.ac.lk}%
  \authorblock{Vijitha Herath}{vijitha@ee.pdn.ac.lk}\hspace*{1.5em}\vspace{-1em}}
\maketitle
\begin{abstract}
Bitemporal remote sensing change detection requires representations that preserve spatial detail while capturing contextual information relevant to changes between acquisitions. Visual test-time training offers an alternative to explicit token-to-token attention by adapting a compact inner model to each image. This letter describes \model, a vanilla Siamese configuration combining the existing hierarchical \vit-Tiny encoder with the multiscale ChangeFormer decoder. The two image streams share learned encoder parameters, while their fast weights are computed independently from their respective spatial features. Each encoder block retains the inherited global simplified SwiGLU and local adaptive depthwise-convolution branches. Temporal comparison is performed by the decoder through learned fusion of paired features at four resolutions, followed by coarse-to-fine aggregation and binary prediction. This separation provides a direct setting for examining image-conditioned inner updates without introducing an additional temporal interaction module. Trained from scratch on LEVIR-CD+ with the same decoder, recipe, and budget as a ChangeFormer control, \model\ reaches an $F_1$ of \FoneTthreeMean$\pm$\FoneTthreeStd\ against \FoneCFMean$\pm$\FoneCFStd\ for ChangeFormer (three seeds each); as in the standard protocol for this dataset, checkpoints are selected on the test split. Trained with both inner updates disabled, the same encoder reaches \FoneTthreeZeroMean$\pm$\FoneTthreeZeroStd. Neither model transfers zero-shot to SYSU-CD, DSIFN-CD, or WHU-CD, and \model\ is the slower of the two at small inputs despite similar parameter and operation counts.
\end{abstract}

\input{figures/architecture/figure}

\section{Introduction}
\IEEEPARstart{B}{uilding} change detection from bitemporal remote sensing imagery supports the monitoring of urban development and the updating of geographic information. Given two co-registered observations, the objective is to distinguish changes in the target structures from variations in appearance that do not alter their change labels. Illumination, shadows, seasonal conditions, and differences in acquisition geometry can complicate this distinction. Accurate dense prediction therefore requires both localized structural information and a representation of the surrounding scene. Building-change benchmarks such as LEVIR-CD and its extension provide a setting for studying this problem~\cite{levir,levirplus}.

Siamese encoders provide a natural starting point by extracting features from each date using shared parameters. Fully convolutional Siamese networks established practical early- and feature-fusion formulations~\cite{fcsn}. Subsequent work has sought richer contextual and object-level representations. For example, DTCDSCN combines change detection with semantic segmentation supervision~\cite{dtcdscn}. Transformer-based approaches offer another route to contextual reasoning. BIT summarizes bitemporal features into a compact set of semantic tokens~\cite{bit}, whereas ChangeFormer combines a hierarchical Transformer encoder with multiscale learned difference features and a decoder~\cite{changeformer}. Semantic change detection, which also labels the land-cover classes involved, has added attention modules and a Dice loss to SCanNet~\cite{11217111} and combined spatio-frequency feature fusion with change-guided attention~\cite{Wijenayake_2026}. These designs illustrate that the representation used before temporal comparison is a central architectural choice.

The cost of spatial context modeling also matters for high-resolution inputs. Full self-attention forms pairwise relations between tokens, although practical architectures can reduce this cost through token reduction or other restrictions. Linear-complexity alternatives have consequently attracted attention in change detection. LSAT uses cross-dimension interactions and structure-aware enhancement~\cite{lsat}; ChangeMamba investigates state-space modeling for spatial and temporal representations~\cite{changemamba}, and precision fusion blocks added to it capture channel-wise temporal variations and per-pixel differences~\cite{wijenayake2026precision}; MambaRefine-CD pairs a MambaVision encoder with region-boundary temporal refinement~\cite{perera2026mambarefinecdmambavisionregionboundarytemporal}; and ChangeRWKV combines a hierarchical RWKV encoder with spatial-temporal fusion~\cite{changerwkv}. Their differing fusion mechanisms mean that an alternative token mixer must be assessed as part of a clearly specified change-detection pipeline, rather than being credited with improvements demonstrated by another architecture. Controlled benchmarks that hold everything but the backbone fixed, and that also report boundary quality and behavior under domain shift, make such an assessment possible~\cite{wasalathilaka2026controlledbenchmarkvisualstatespace}.

Test-time training (TTT) was introduced to adapt a model to each test input through a self-supervised auxiliary task~\cite{sun2020ttt,gandelsman2022tttmae}, alongside fully test-time adaptation methods that update a deployed model without labels~\cite{wang2021tent,liang2024ttasurvey}. TTT layers instead build the adaptation into the architecture: they represent context through the parameters of a small inner model that is updated using the input sequence~\cite{ttt}, a view closely related to fast-weight programmers~\cite{schlag2021fastweights} and since extended to large-chunk updates~\cite{zhang2025testtimetrainingright} and learned test-time memory~\cite{behrouz2024titans}, which ChangeTitans has applied to change detection~\cite{changetitans}. The visual \vit\ architecture adapts this principle to images through full-batch spatial updates and complementary global and local inner models~\cite{vit3}. Unlike a fixed feature transformation, the effective inner weights depend on the current image. However, per-image adaptation does not itself compare acquisitions: independently adapted date features must still be combined to produce a change map. Whether this representation is beneficial for building-change detection is therefore an empirical question.

This letter develops a deliberately unmodified encoder--decoder configuration for examining that question. \model\ uses a shared hierarchical \vit-Tiny encoder and the existing ChangeFormer V6 decoder. Each date generates its own fast weights; no temporal exchange is inserted into the encoder. The method description distinguishes these image-conditioned weights from shared learned parameters and identifies the decoder as the point of temporal fusion. The accompanying evaluation structure isolates the global and local inner updates while retaining the decoder. These are an architectural adaptation and an experimental study design, rather than new inner models or a new temporal-fusion mechanism. On LEVIR-CD+, the encoder with active inner updates exceeds a ChangeFormer control that shares its decoder, recipe, and budget, whereas the same encoder trained without updates does not (Section~\ref{sec:experiments}). That advantage does not carry over to zero-shot evaluation on other datasets, where every model trained only on LEVIR-CD+ fails (Section~\ref{sec:crossdomain}).

\section{Method}
\subsection{Siamese Multiscale Representation}
Let $I_A,I_B\in\mathbb{R}^{3\times H\times W}$ be a co-registered image pair and $Y\in\{0,1\}^{H\times W}$ its binary change mask. A single encoder $E_\theta$ is applied separately to both images:
\begin{equation}
 \{F_t^i\}_{i=1}^{4}=E_\theta(I_t),\qquad t\in\{A,B\}.
 \label{eq:encoder}
\end{equation}
For dimensions divisible by 32, $F_t^i\in\mathbb{R}^{C_i\times H/s_i\times W/s_i}$, where $s_i=2^{i+1}$. The stage widths are $(64,128,320,512)$, the block depths are $(1,3,9,4)$, and the global head counts are $(2,4,10,16)$. Thus every global head has width $d=32$. A convolutional stem and successive patch-merging operations reduce the spatial resolution; stage-specific layer normalization produces the four decoder inputs. The diagram in Fig.~\ref{fig:architecture} separates shared parameterization from independent date processing.

In each block, a residual $3\times3$ depthwise convolution first provides conditional positional information. A pre-normalized TTT module and a pre-normalized spatial feed-forward module then operate through residual connections. Suppressing stochastic-depth operators for readability, the block is
\begin{align}
 X_p&=X+\dw_{\mathrm{pos}}(X),\nonumber\\
 X_m&=X_p+\operatorname{TTT}(\LN(X_p)),\nonumber\\
 X_{\mathrm{out}}&=X_m+\operatorname{MLP}_{\mathrm{sp}}(\LN(X_m)).
 \label{eq:block}
\end{align}
Spatial reshaping is implicit. The spatial MLP contains channel projections and an intermediate residual depthwise convolution. These fixed learned convolutions are distinct from the adaptive local convolution inside the TTT module.

\subsection{Independent Image-Conditioned Inner Updates}
At a particular block, let $X\in\mathbb{R}^{N\times C}$ denote normalized tokens from one image, with $N=H_iW_i$. A learned projection produces global $(Q_g,K_g,V_g)$ with $C$ channels each and local $(Q_\ell,K_\ell,V_\ell)$ with $d$ channels each. Two-dimensional rotary positional encoding is applied to global queries and keys before head-wise processing. The global tensors are arranged into $h$ heads of width $d=C/h$, while local tensors retain their two-dimensional spatial organization.

For one global head, the inherited simplified SwiGLU inner model is
\begin{equation}
 f_{W_g}(U)=(UW_1)\odot\silu(UW_2),
 \quad W_g=(W_1,W_2),
 \label{eq:glu}
\end{equation}
where $W_1,W_2\in\mathbb{R}^{d\times d}$ and $\odot$ denotes elementwise multiplication. This inner model has no additional output matrix. In parallel, the local inner model applies a depthwise convolution,
\begin{equation}
 f_{W_\ell}(U)=\dw_{3\times3}(U;W_\ell),
 \label{eq:local}
\end{equation}
with one spatial kernel per local channel and padding that preserves resolution.

Both branches use the inherited negative dot-product objective~\cite{vit3}. For a global head it is
\begin{equation}
 \mathcal{L}_{g}(W_g;K_g,V_g)
 =-\frac{s}{N}\sum_{n=1}^{N}
 \left\langle f_{W_g}(K_g)_n,V_{g,n}\right\rangle.
 \label{eq:innerloss}
\end{equation}
For local channel $c$, the corresponding loss is $-s/N$ times the sum of $f_{W_\ell}(K_\ell)_{c,u,v}V_{\ell,c,u,v}$ over spatial positions. Each local kernel is updated from its own channel. The implementation uses the inherited scale $s=9^{-1/2}$ in both branches; the value 9 is the kernel area, not the global head width. The inner targets are learned value projections, not change-mask labels.

For branch $b\in\{g,\ell\}$ and image $t$, one update produces
\begin{equation}
 W_{t,b}^{\star}=W_{0,b}
 -\eta_b\mathcal{R}_{b}\!\left(
 \nabla_W\mathcal{L}_{t,b}(W_{0,b})\right).
 \label{eq:update}
\end{equation}
Here $W_{0,b}$ is a learned initialization and $\mathcal{R}_b$ is the gradient rescaling used by the implementation. Each column vector $g$ of a global matrix gradient is transformed into $g/(1+\|g\|_2)$. Each local kernel gradient $G$ is transformed into $G/(1+\|G\|_F)$. Equation~\eqref{eq:update} therefore does not represent an unmodified gradient-descent step. The vanilla code uses $\eta_g=\eta_\ell=1$. These are inner architectural step sizes, distinct from the outer optimizer learning rate.

The update uses all spatial positions of the current image. It neither pools keys and values across training examples nor combines dates. The fast weights for $A$ and $B$ start from the same learned initialization but are computed independently. They are temporary tensors and do not replace the persistent model parameters during inference. The module then evaluates the queries:
\begin{equation}
 O=P_o\!\left(\cat\left[
 f_{W_{t,g}^{\star}}(Q_g),
 f_{W_{t,\ell}^{\star}}(Q_\ell)\right]\right).
 \label{eq:output}
\end{equation}
The global heads are reassembled and the local map is flattened before concatenation. Consequently, $P_o$ maps $C+d$ channels back to $C$.

The gradients in~\eqref{eq:update} are computed analytically with differentiable tensor operations. Outer supervised training can differentiate through these operations, while inference does not require an outer optimizer step or ground-truth labels. The global branch can condition an output location on all image tokens through its fast weights. The local branch applies a spatially local query transformation, although its adaptive kernel is itself conditioned on the whole spatial map. This distinction is important: ``local'' describes the convolutional query operator, not an exclusively local source of adaptation.

\subsection{Learned Temporal Decoding and Supervision}
The ChangeFormer V6 decoder~\cite{changeformer} receives the paired pyramids. At scale $i$, a shared linear projection $P_i$ maps both dates to a common decoder width $D$:
\begin{equation}
 A_i=P_i(F_A^i),\qquad B_i=P_i(F_B^i).
 \label{eq:project}
\end{equation}
All runs use $D=256$. A learned convolutional difference block $\phi_i$ transforms the ordered concatenation $[A_i,B_i]$. Coarse-to-fine decoding is
\begin{align}
 D_4&=\phi_4(\cat[A_4,B_4]),\nonumber\\
 D_i&=\phi_i(\cat[A_i,B_i])+\up_{i+1\rightarrow i}(D_{i+1}),
 \quad i=3,2,1.
 \label{eq:decoder}
\end{align}
The interpolation operator $\up$ uses bilinear resizing. Each difference block contains convolution, activation, and normalization operations. Thus the decoder learns how to compare temporal features; it does not simply compute an absolute difference. Likewise, shared encoder weights alone do not guarantee exact invariance of the complete prediction to swapping the input dates.

All $D_i$ are aligned to the finest decoder grid and fused:
\begin{equation}
 D_f=\psi\!\left(\cat[
 \up_{4\rightarrow1}(D_4),\up_{3\rightarrow1}(D_3),
 \up_{2\rightarrow1}(D_2),D_1]\right),
 \label{eq:fusion}
\end{equation}
where $\psi$ is a $1\times1$ convolution followed by batch normalization. Two transposed-convolution upsampling stages, each followed by residual refinement, restore the spatial resolution. A final convolution predicts two logits per pixel. The integration restores the original input extent by cropping padded outputs when possible, or by interpolation otherwise. The final mask is obtained by the class-wise argmax of the final logits; any use of multiscale inference must be documented separately.

Four auxiliary heads produce predictions at the decoder scales. With deep supervision enabled, the outer objective is
\begin{equation}
 \mathcal{L}_{\mathrm{sup}}=
 \sum_{j=1}^{5}\lambda_j\,
 \CE(Z_j,\mathcal{S}_j(Y)),
 \label{eq:outerloss}
\end{equation}
where $(Z_1,\ldots,Z_5)$ correspond to scales $(4,3,2,1)$ and the final output. The operator $\mathcal{S}_j$ resizes labels to the prediction grid using nearest-neighbor interpolation; cross-entropy consumes raw two-class logits. The coefficients are $(\lambda_1,\ldots,\lambda_5)=(\DSWeights)$. This outer loss jointly trains the encoder initialization, feature projections, and decoder. It is separate from the unlabeled inner objective in~\eqref{eq:innerloss}.

\subsection{Computational Scope}
For fixed channel dimensions and one inner update, the global inner-model operations scale as $O(Nhd^2)$ and the local kernel operations as $O(Ndk^2)$, with $k=3$. Channel projections and the spatial feed-forward operations are also linear in $N$ at fixed widths. Applying the shared encoder twice doubles its per-image work without doubling its learned parameters. This inherited token-linear structure avoids constructing a full $N\times N$ attention matrix, but it does not by itself establish lower measured latency, memory use, or parameter count than a competing detector. End-to-end profiling must include both date passes, inner-update arithmetic, and decoding.

\section{Experiments}\label{sec:experiments}
\subsection{Dataset and Evaluation Protocol}
The benchmark is LEVIR-CD+, the extended building-change dataset linked by the LEVIR authors~\cite{levir,levirplus}, in its \PatchSize$\times$\PatchSize\ patch release: each 1024$\times$1024 scene pair is cut into 16 non-overlapping patches with stride \PatchSize. The \NTrainParents\ training scenes yield \NTrain\ training patches and the other \NTestParents\ scenes yield \NTest\ test patches, so training and test scenes are disjoint. In this release, as in the literature that uses it, the validation set is a byte-identical copy of the test set (identical file lists and mask checksums). Every score reported here is therefore a validation score in the strict sense, and no independent test estimate exists. Images are normalized with ImageNet statistics, masks are binarized at 127, and no data augmentation is applied.

Change-class precision, recall, F1, and intersection over union (IoU) will summarize pixel-level predictions. With true positives $T_P$, false positives $F_P$, and false negatives $F_N$, their definitions are
\begin{align}
 P&=\frac{T_P}{T_P+F_P},&
 R&=\frac{T_P}{T_P+F_N},\nonumber\\
 F_1&=\frac{2T_P}{2T_P+F_P+F_N},&
 \mathrm{IoU}&=\frac{T_P}{T_P+F_P+F_N}.
\end{align}
The counts are accumulated over every pixel of all \NTest\ test patches before the ratios are formed (dataset-level aggregation). No pixels are ignored, and a constant $\epsilon=10^{-8}$ is added to each denominator. Evaluation runs on the \PatchSize$\times$\PatchSize\ patches; scenes are not reconstructed. We also report boundary $F_1$ (BF1) with a tolerance of \BndTol\ pixels.

\subsection{Implementation and Comparisons}
Table~\ref{tab:setup} lists the training and evaluation settings; every run shares them, and runs differ only in the encoder, its initialization, the inner step sizes, or the seed. The deep-supervision weights apply from the coarsest output to the full-resolution head. ChangeFormer, \model, and \model\ without inner updates are each trained with two seeds beyond seed \Seed\ to measure run-to-run spread. \model-P initializes the encoder from the ImageNet-pretrained H-\vit-T weights released with~\cite{vit3}, once without and once with layer-wise learning-rate decay. ChangeFormer-P initializes its encoder from SegFormer MiT-B2 pretrained on ADE20K, the source the ChangeFormer authors used; since their converted file is no longer available, the released SegFormer weights are mapped by name, and the three $3\times3$ patch-embedding kernels are embedded at the centre of the encoder's $7\times7$ kernels, which reproduces their output exactly. In every pretrained run only the encoder is initialized; the decoder is trained from scratch.

\begin{table}[t]
\caption{Training and evaluation settings shared by all runs.}
\label{tab:setup}
\centering\footnotesize
\renewcommand{\arraystretch}{1.08}
\begin{tabular}{@{}>{\raggedright\arraybackslash}p{0.30\columnwidth}>{\raggedright\arraybackslash}p{0.64\columnwidth}@{}}
\toprule
\multicolumn{2}{@{}l}{\textit{Data}}\\
Dataset & LEVIR-CD+, \PatchSize$\times$\PatchSize\ patches, stride \PatchSize\\
Patches & \NTrain\ train / \NTest\ test (val $=$ test)\\
Scenes & \NTrainParents\ train / \NTestParents\ test, disjoint\\
Preprocessing & ImageNet normalization; masks binarized at 127; no augmentation\\
\midrule
\multicolumn{2}{@{}l}{\textit{Optimization}}\\
Optimizer & AdamW ($\beta_1{=}0.9$, $\beta_2{=}0.999$), one parameter group\\
Learning rate & $\LR$, linear per-iteration decay to 0\\
Weight decay & \WD\\
Budget & \Iters\ iterations, batch size \BatchSize\\
Loss & Cross-entropy on 5 outputs, weights (\DSWeights)\\
Precision & fp32, deterministic cuDNN\\
\midrule
\multicolumn{2}{@{}l}{\textit{Model}}\\
Decoder & ChangeFormer V6, width 256\\
Stochastic depth & 0.2 (\model); 0.1 (ChangeFormer encoder)\\
Initialization & Scratch; \model-P: ImageNet H-\vit-T, layer decay 0.8 (one run); ChangeFormer-P: ADE20K SegFormer MiT-B2\\
Seeds & \Seed; ChangeFormer, \model, and \model\ w/o updates also 1 and 2\\
\midrule
\multicolumn{2}{@{}l}{\textit{Selection and evaluation}}\\
Selection & Best $F_1$, evaluated every \EvalEvery\ iterations; no early stopping\\
Inference & Argmax of the final head, single scale, no TTA\\
Metrics & Dataset-level $P$, $R$, $F_1$, IoU; BF1 at \BndTol\,px\\
Hardware & NVIDIA RTX 6000 Ada\\
\bottomrule
\end{tabular}
\end{table}

\begin{table*}[t]
\caption{LEVIR-CD+ results (\%). ChangeFormer and \model\ share decoder, recipe, and budget. Scores are on the test split, which also served for checkpoint selection (test-as-val protocol). Seed rows give mean$\pm$standard deviation. Pretrained rows differ in pretraining data as well as encoder.}
\label{tab:comparison}
\centering\footnotesize
\begin{tabular*}{\textwidth}{@{\extracolsep{\fill}}llcccccc@{}}
\toprule
Method & Init. & Params (M) & $P$ & $R$ & $F_1$ & IoU & BF1\\
\midrule
ChangeFormer (3 seeds) & scratch & 41.03 & 77.03$\pm$1.17 & 80.76$\pm$0.83 & 78.84$\pm$0.21 & 65.07$\pm$0.29 & 70.58$\pm$0.21 \\
\model (3 seeds) & scratch & 40.28 & 80.54$\pm$1.86 & 80.70$\pm$0.68 & 80.61$\pm$0.90 & 67.52$\pm$1.26 & 72.69$\pm$0.67 \\
\midrule
ChangeFormer-P & ADE20K & 41.03 & 84.01 & 83.95 & 83.98 & 72.38 & 79.30 \\
\model-P & ImageNet & 40.28 & 86.65 & 83.90 & 85.25 & 74.29 & 78.24 \\
\model-P + layer decay & ImageNet & 40.28 & 87.34 & 83.38 & 85.31 & 74.39 & 78.47 \\
\bottomrule
\end{tabular*}

\end{table*}

ChangeFormer, the only baseline, is trained under exactly this protocol: it shares the decoder, loss, optimizer, schedule, and budget of \model, so the two differ only in the encoder (Table~\ref{tab:comparison}). It is our run of the ChangeFormer encoder~\cite{changeformer} under this recipe, not the score reported by its authors. Over three seeds each, \model\ reaches \FoneTthreeMean$\pm$\FoneTthreeStd\ $F_1$ and ChangeFormer \FoneCFMean$\pm$\FoneCFStd, a margin of \DeltaMeanMeanCF\ points. ChangeFormer's seed spread is the smaller of the two, and the margin exceeds either spread. The two models have similar sizes, \ParamsTthree\,M and \ParamsCF\,M parameters. Pretraining helps both encoders: ChangeFormer-P reaches \FoneCFPre\ and \model-P \FonePre--\FonePreLD. These rows are listed separately, and they are not a controlled comparison of encoders: the two start from different data and tasks (ADE20K segmentation versus ImageNet classification), and each is a single seed.

\subsection{Controlled Inner-Update Ablations}
Table~\ref{tab:ablation} defines a planned factorial study of the global and local inner updates. Setting a step size to zero retains that branch and its learned initialization; it does not remove the branch. Independently training each configuration tests adaptation within the learning system, whereas changing a step size only at inference probes reliance of a particular checkpoint on its updates. These protocols will be reported separately; Table~\ref{tab:ablation} covers the first.

All four configurations use seed \Seed\ and are otherwise identical. Training with both updates gives an $F_1$ of \FoneTthree, the global update alone \FoneGlobalOnly, the local update alone \FoneLocalOnly, and neither \FoneTthreeZero. Relative to no update, the global update alone is worth \DeltaGlobalNone\ points, the local update alone \DeltaLocalNone, and both together \DeltaFullNone. Seed \Seed\ is the strongest of the three \model\ seeds, so the seed means are the fairer comparison: \FoneTthreeMean$\pm$\FoneTthreeStd\ with both updates against \FoneTthreeZeroMean$\pm$\FoneTthreeZeroStd\ without, a gap of \DeltaMeanMeanNone\ points that is well outside either spread. Either update alone already recovers most of that gap. The single-branch runs have one seed each and fall within the full model's seed spread, so these data do not separate the two branches. Without updates, \model\ scores at the level of ChangeFormer (\FoneTthreeZeroMean\ against \FoneCFMean).

\begin{table}[t]
\caption{Training-time inner-update ablations on LEVIR-CD+ (\%). The upper block is seed \Seed; the lower block gives mean$\pm$standard deviation over seeds. A zero step size keeps the branch and its learned initialization but applies no update. Best iter.\ is the selected checkpoint.}
\label{tab:ablation}
\centering\footnotesize\setlength{\tabcolsep}{3pt}
\begin{tabular}{@{}lccccc@{}}
\toprule
Configuration & $\eta_g$ & $\eta_\ell$ & $F_1$ & IoU & Best iter.\\
\midrule
Both (vanilla) & 1 & 1 & 81.64 & 68.97 & 43k \\
Global only & 1 & 0 & 80.18 & 66.91 & 48k \\
Local only & 0 & 1 & 80.80 & 67.79 & 28k \\
Neither & 0 & 0 & 78.73 & 64.92 & 38k \\
\midrule
Both (3 seeds) & 1 & 1 & 80.61$\pm$0.90 & 67.52$\pm$1.26 & -- \\
Neither (3 seeds) & 0 & 0 & 78.63$\pm$0.20 & 64.79$\pm$0.27 & -- \\
\bottomrule
\end{tabular}

\end{table}

\subsection{Qualitative Evidence}
\begin{figure*}[!t]
  \centering
  \includegraphics[width=\textwidth,
                   height=\dimexpr\textheight-9\baselineskip\relax,
                   keepaspectratio]{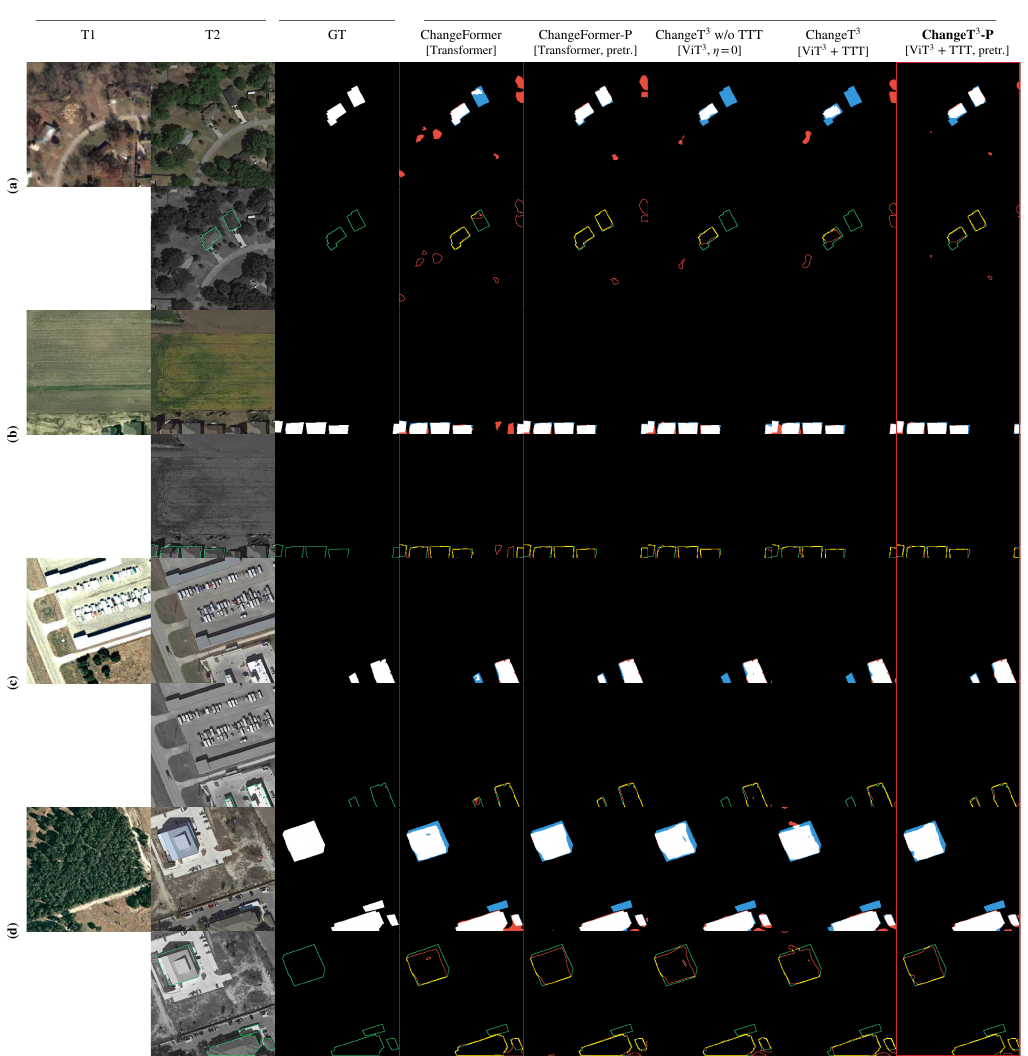}
  \par\smallskip
  \definecolor{qualFP}{HTML}{E74C3C}\definecolor{qualFN}{HTML}{3498DB}%
  \definecolor{qualGT}{HTML}{2ECC71}\definecolor{qualOV}{HTML}{FFE600}%
  \providecommand{\qualswatch}[1]{{\setlength{\fboxsep}{0pt}\fcolorbox{black!50}{#1}{\rule{0pt}{5pt}\rule{5pt}{0pt}}}}%
  \providecommand{\qualline}[1]{\textcolor{#1}{\rule[0.45ex]{11pt}{1.4pt}}}%
  {\footnotesize
  \textbf{Change map:}~\qualswatch{white}~TP\quad\qualswatch{black}~TN\quad
  \qualswatch{qualFP}~FP\quad\qualswatch{qualFN}~FN\qquad
  \textbf{Boundary:}~\qualline{qualGT}~reference\quad\qualline{qualFP}~prediction\quad
  \qualline{qualOV}~overlap\par}
  \caption{Qualitative comparison on the LEVIR-CD+ test split. Samples (a)--(d) are
  the four highest-ranked test patches from distinct scenes under the selection rule
  of Section~\ref{sec:experiments}. For each sample, the upper row shows the change
  prediction as an error map, where white, black, red and blue denote true positive,
  true negative, false positive and false negative pixels. The lower row shows
  structural agreement between the reference contour (green) and the predicted
  contour (red), with yellow indicating overlap. ChangeFormer and \model\ share the
  same decoder and training recipe; they differ only in the encoder and in whether
  its per-image test-time update is active. The -P variants initialize the encoder
  from pretrained weights (ADE20K for ChangeFormer, ImageNet for \model).}
  \label{fig:qual_levir}
\end{figure*}

Figure~\ref{fig:qual_levir} shows four LEVIR-CD+ test patches chosen by a fixed rule from the first 100 test images: a change fraction between 2\% and 40\%, at least one changed component of 64 pixels or more, an $F_1$ of at least 0.5 for the better \model\ variant, and, among those, the largest $F_1$ margin over ChangeFormer, taking at most one patch per source scene. Because the rule selects for cases where \model\ is ahead, the figure illustrates error types rather than measuring an advantage. All predictions are read from the cached outputs of the evaluation, not re-run for the figure.

\subsection{Zero-Shot Cross-Domain Evaluation}\label{sec:crossdomain}
Following domain-shift analyses of remote-sensing backbones~\cite{wasalathilaka2026controlledbenchmarkvisualstatespace}, every model trained on LEVIR-CD+ is also evaluated, without any tuning, on the test splits of SYSU-CD (\NSYSU\ pairs), DSIFN-CD (\NDSIFN), and WHU-CD (\NWHU). The checkpoint is the one selected on LEVIR-CD+ and the decision is the argmax of the two-class logits; nothing is selected or calibrated on a target. SYSU-CD and WHU-CD pairs are 256$\times$256 and used as they are. DSIFN-CD pairs are 512$\times$512; they are cut into four non-overlapping 256$\times$256 tiles at native resolution, and the stitched predictions are scored on the full images. Resizing them to 256$\times$256 instead, as earlier code bases do, lowers every model's score and leaves the conclusions unchanged.

\begin{table*}[t]
\caption{Zero-shot cross-domain evaluation (\%) of models trained only on LEVIR-CD+. Seed rows give mean$\pm$standard deviation. Cross-domain mean $F_1$ averages the three targets; drop is in-domain $F_1$ minus that mean.}
\label{tab:crossdomain}
\centering\scriptsize\setlength{\tabcolsep}{3pt}
\begin{tabular*}{\textwidth}{@{\extracolsep{\fill}}lcccccccccc@{}}
\toprule
 & \multicolumn{2}{c}{LEVIR-CD+ (in-domain)} & \multicolumn{2}{c}{SYSU-CD} & \multicolumn{2}{c}{DSIFN-CD} & \multicolumn{2}{c}{WHU-CD} & \multicolumn{2}{c}{Cross-domain}\\
\cmidrule(lr){2-3}\cmidrule(lr){4-5}\cmidrule(lr){6-7}\cmidrule(lr){8-9}\cmidrule(l){10-11}
Model & $F_1$ & IoU & $F_1$ & IoU & $F_1$ & IoU & $F_1$ & IoU & mean $F_1$ & drop\\
\midrule
ChangeFormer (3 seeds) & 78.84$\pm$0.21 & 65.07$\pm$0.29 & 2.23$\pm$0.32 & 1.13$\pm$0.17 & 15.86$\pm$1.33 & 8.62$\pm$0.79 & 1.84$\pm$1.08 & 0.93$\pm$0.55 & 6.64$\pm$0.40 & 72.20$\pm$0.62 \\
\model\ w/o TTT (3 seeds) & 78.63$\pm$0.20 & 64.79$\pm$0.27 & 0.80$\pm$0.21 & 0.40$\pm$0.10 & 7.97$\pm$0.62 & 4.15$\pm$0.34 & 1.92$\pm$0.88 & 0.97$\pm$0.45 & 3.57$\pm$0.22 & 75.07$\pm$0.38 \\
\model (3 seeds) & 80.61$\pm$0.90 & 67.52$\pm$1.26 & 0.87$\pm$0.36 & 0.44$\pm$0.18 & 3.41$\pm$0.73 & 1.73$\pm$0.38 & 1.22$\pm$0.21 & 0.61$\pm$0.11 & 1.83$\pm$0.08 & 78.78$\pm$0.96 \\
\midrule
ChangeFormer-P & 83.98 & 72.38 & 3.52 & 1.79 & 18.13 & 9.97 & 6.04 & 3.12 & 9.23 & 74.75 \\
\model-P & 85.25 & 74.29 & 3.86 & 1.97 & 11.83 & 6.29 & 23.53 & 13.33 & 13.07 & 72.18 \\
\model-P + l.d. & 85.31 & 74.39 & 2.29 & 1.16 & 8.97 & 4.70 & 30.45 & 17.96 & 13.90 & 71.41 \\
\bottomrule
\end{tabular*}

\end{table*}

\begin{table*}[t]
\caption{Cost of one forward pass on an image pair at 256$\times$256, fp32, on one \EffGPU\ GPU. Latency is mean$\pm$standard deviation over 100 timed passes. $^\dagger$Inner gradients not computed; the output is identical to $\eta{=}0$. The training columns come from the training logs on shared GPUs; contention differed between runs, so they are indicative only and not comparable across rows.}
\label{tab:efficiency}
\centering\footnotesize
\begin{tabular*}{\textwidth}{@{\extracolsep{\fill}}lccccccc@{}}
\toprule
Model & Params (M) & GFLOPs & Latency (ms) & Throughput & Peak mem (GB) & Train & Train mem\\
 & & & $b{=}1$ & pairs/s, $b{=}8$ & $b{=}1$ / $b{=}8$ & it/s & (GB)\\
\midrule
ChangeFormer & 41.03 & 277.5 & 16.2$\pm$1.3 & 79.1 & 0.45 / 2.96 & 2.81 & 7.11 \\
\model & 40.28 & 284.4 & 89.9$\pm$3.1 & 50.4 & 0.60 / 3.12 & 1.53 & 9.55 \\
\model, $\eta{=}0$ & 40.28 & 284.4 & 92.2$\pm$9.0 & 49.7 & 0.75 / 3.27 & 2.33 & 9.55 \\
\model, update skipped$^\dagger$ & -- & -- & 60.2$\pm$3.1 & 61.9 & 0.60 / 3.12 & -- & -- \\
\bottomrule
\end{tabular*}

\end{table*}

Table~\ref{tab:crossdomain} reports the scores. Every model trained from scratch is near the floor on every target: the mean $F_1$ over the three targets is \XDCF\ for ChangeFormer, \XDTthreeZero\ for \model\ without updates, and \XDTthree\ for \model, so the inner update that helps in-domain does not help here, and \model\ is the lowest of the three. The failure is one of recall. On SYSU-CD, \model\ marks \SYSUPredFracTthree\% of the pixels as changed against \SYSUGTFrac\% in the reference, with a precision of \SYSUPrecTthree\ and a recall of \SYSURecTthree; DSIFN-CD and WHU-CD show the same pattern. Differences in what each dataset labels as change may contribute on SYSU-CD and DSIFN-CD, whose labels cover more than buildings, but WHU-CD labels building changes as LEVIR-CD+ does and shows the same collapse, so label definitions alone do not explain it. Pretraining helps most: the cross-domain mean reaches \XDPreLD\ for \model-P with layer decay and \XDCFPre\ for ChangeFormer-P, mainly through WHU-CD. These are single-source, zero-shot results with no adaptation, which is the setting a test-time adaptation method would have to address~\cite{wang2021tent,liang2024ttasurvey}.

\subsection{Efficiency}
Table~\ref{tab:efficiency} compares cost on one \EffGPU\ GPU in fp32, over 20 warm-up and 100 timed forward passes of an image pair with synchronization. The two models have nearly the same parameter counts and operation counts (\GFLOPsTthree\ and \GFLOPsCF\ GFLOPs at 256$\times$256, counting the analytical inner-update products), and similar peak memory. Their latency is not similar: at batch size 1 and 256$\times$256, \model\ takes \LatTthree\,ms against \LatCF\,ms, and at batch size 8 it processes \ThrTthree\ pairs per second against \ThrCF. Skipping the inner-gradient computation, which leaves the output equal to $\eta{=}0$, shows that the update adds \InnerLatPct\% to the batch-1 latency and \InnerLatPctBEight\% to the batch-8 time, with no measurable extra memory. The gap narrows with input size (Table~\ref{tab:scaling}), from \ScaleRatioLow$\times$ at 256$\times$256 to \ScaleRatioHigh$\times$ at 1024$\times$1024, while operations and memory stay close: at small inputs the cost is dominated by the number of operations launched rather than by their arithmetic. Timings on this shared machine also drifted within a profiling run, as identical architectures measured at different points differed by up to \DriftLatX$\times$ in batch-1 latency and \DriftThrPct\% in batch-8 throughput. The latencies are therefore indicative, and the parameter, operation, and memory counts are the reliable comparison.


\begin{table}[t]
\caption{Batch-1 latency, peak allocated memory, and operations against input size.}
\label{tab:scaling}
\centering\footnotesize\setlength{\tabcolsep}{4pt}
\begin{tabular}{@{}rcccccc@{}}
\toprule
 & \multicolumn{3}{c}{ChangeFormer} & \multicolumn{3}{c}{\model}\\
\cmidrule(lr){2-4}\cmidrule(l){5-7}
Input & ms & GB & GFLOPs & ms & GB & GFLOPs\\
\midrule
$256^2$ & 17.3 & 0.45 & 277 & 88.3 & 0.60 & 284 \\
$384^2$ & 29.2 & 0.81 & 627 & 103.7 & 0.96 & 640 \\
$512^2$ & 48.1 & 1.31 & 1120 & 123.0 & 1.46 & 1138 \\
$768^2$ & 110.4 & 2.74 & 2556 & 197.2 & 2.89 & 2560 \\
$1024^2$ & 208.6 & 4.75 & 4637 & 345.7 & 4.90 & 4551 \\
\bottomrule
\end{tabular}

\end{table}

\section{Conclusion}
This letter specifies a vanilla adaptation of hierarchical visual TTT to bitemporal change detection using an existing multiscale ChangeFormer decoder. Its defining separation is between shared learned encoder parameters, independent image-conditioned fast weights, and decoder-based temporal comparison. The architecture provides a setting for evaluating the contribution of the inherited global and local inner updates without adding temporal modules. On LEVIR-CD+, with the decoder, recipe, and budget held fixed, the \vit\ encoder with active inner updates reaches an $F_1$ of \FoneTthreeMean$\pm$\FoneTthreeStd\ against \FoneCFMean$\pm$\FoneCFStd\ for the ChangeFormer encoder over three seeds each, while the same encoder trained with both updates disabled reaches \FoneTthreeZeroMean$\pm$\FoneTthreeZeroStd, level with ChangeFormer. The advantage stays in-domain: evaluated zero-shot on SYSU-CD, DSIFN-CD, and WHU-CD, every scratch model fails through missed changes, \model\ most of all, and pretraining helps where the inner update does not. At similar parameter and operation counts, \model\ is slower than ChangeFormer at small inputs, and the difference narrows at larger ones. These results rest on one training dataset, one baseline, and checkpoints selected on the test split; the controls are correlational evidence about the inner updates, not a causal account. Adapting the fast weights to the target domain at test time~\cite{sun2020ttt,wang2021tent}, which this architecture permits but the present study does not attempt, is the natural next step.

\bibliographystyle{IEEEtran}
\bibliography{references}

\begin{thebibliography}{10}
\providecommand{\url}[1]{#1}
\csname url@samestyle\endcsname
\providecommand{\newblock}{\relax}
\providecommand{\bibinfo}[2]{#2}
\providecommand{\BIBentrySTDinterwordspacing}{\spaceskip=0pt\relax}
\providecommand{\BIBentryALTinterwordstretchfactor}{4}
\providecommand{\BIBentryALTinterwordspacing}{\spaceskip=\fontdimen2\font plus
\BIBentryALTinterwordstretchfactor\fontdimen3\font minus \fontdimen4\font\relax}
\providecommand{\BIBforeignlanguage}[2]{{%
\expandafter\ifx\csname l@#1\endcsname\relax
\typeout{** WARNING: IEEEtran.bst: No hyphenation pattern has been}%
\typeout{** loaded for the language `#1'. Using the pattern for}%
\typeout{** the default language instead.}%
\else
\language=\csname l@#1\endcsname
\fi
#2}}
\providecommand{\BIBdecl}{\relax}
\BIBdecl

\bibitem{levir}
H.~Chen and Z.~Shi, ``A spatial-temporal attention-based method and a new dataset for remote sensing image change detection,'' \emph{Remote Sensing}, vol.~12, no.~10, p. 1662, 2020.

\bibitem{levirplus}
\BIBentryALTinterwordspacing
{S2Looking dataset maintainers}, ``{LEVIR-CD+} dataset,'' Online dataset repository, accessed: September 24, 2026. [Online]. Available: \url{https://github.com/S2Looking/Dataset}
\BIBentrySTDinterwordspacing

\bibitem{fcsn}
R.~C. Daudt, B.~L. Saux, and A.~Boulch, ``Fully convolutional {Siamese} networks for change detection,'' in \emph{IEEE International Conference on Image Processing (ICIP)}, 2018, open version: arXiv:1810.08462.

\bibitem{dtcdscn}
Y.~Liu, C.~Pang, Z.~Zhan, X.~Zhang, and X.~Yang, ``Building change detection for remote sensing images using a dual-task constrained deep {Siamese} convolutional network model,'' \emph{IEEE Geoscience and Remote Sensing Letters}, vol.~18, pp. 811--815, 2021.

\bibitem{bit}
H.~Chen, Z.~Qi, and Z.~Shi, ``Remote sensing image change detection with {Transformers},'' \emph{IEEE Transactions on Geoscience and Remote Sensing}, 2021, online publication; open version: arXiv:2103.00208.

\bibitem{changeformer}
W.~G.~C. Bandara and V.~M. Patel, ``A {Transformer}-based {Siamese} network for change detection,'' in \emph{IEEE International Geoscience and Remote Sensing Symposium (IGARSS)}, 2022, open version: arXiv:2201.01293.

\bibitem{11217111}
R.~M. A. M.~B. Ratnayake, W.~M. B. S.~K. Wijenayake, D.~M. U.~P. Sumanasekara, G.~M. R.~I. Godaliyadda, H.~M. V.~R. Herath, and M.~P.~B. Ekanayake, ``Enhanced {SCanNet} with {CBAM} and dice loss for semantic change detection,'' in \emph{2025 Moratuwa Engineering Research Conference (MERCon)}, 2025, pp. 84--89.

\bibitem{Wijenayake_2026}
\BIBentryALTinterwordspacing
B.~Wijenayake, A.~Ratnayake, P.~Sumanasekara, R.~Godaliyadda, P.~Ekanayake, V.~Herath, and N.~Wasalathilaka, ``{Mamba-FCS}: Joint spatio-frequency feature fusion, change-guided attention, and {SeK} inspired loss for enhanced semantic change detection in remote sensing,'' \emph{IEEE Journal of Selected Topics in Applied Earth Observations and Remote Sensing}, vol.~19, pp. 7680--7698, 2026. [Online]. Available: \url{http://dx.doi.org/10.1109/JSTARS.2026.3663066}
\BIBentrySTDinterwordspacing

\bibitem{lsat}
T.~Lei, Y.~Xu, H.~Ning, Z.~Lv, C.~Min, Y.~Jin, and A.~K. Nandi, ``Lightweight structure-aware {Transformer} network for remote sensing image change detection,'' \emph{IEEE Geoscience and Remote Sensing Letters}, vol.~21, pp. 1--5, 2024, article 6000305.

\bibitem{changemamba}
H.~Chen, J.~Song, C.~Han, J.~Xia, and N.~Yokoya, ``{ChangeMamba}: Remote sensing change detection with spatiotemporal state space model,'' \emph{arXiv preprint arXiv:2404.03425}, 2024.

\bibitem{wijenayake2026precision}
\BIBentryALTinterwordspacing
W.~M. B. S.~K. Wijenayake, R.~M. A. M.~B. Ratnayake, D.~M. U.~P. Sumanasekara, N.~S. Wasalathilaka, M.~Piratheepan, G.~M. R.~I. Godaliyadda, M.~P.~B. Ekanayake, and H.~M. V.~R. Herath, ``Precision spatio-temporal feature fusion for robust remote sensing change detection,'' in \emph{2025 IEEE 19th International Conference on Industrial and Information Systems (ICIIS)}.\hskip 1em plus 0.5em minus 0.4em\relax IEEE, Jan. 2026, pp. 557--562. [Online]. Available: \url{http://dx.doi.org/10.1109/ICIIS69028.2026.11450773}
\BIBentrySTDinterwordspacing

\bibitem{perera2026mambarefinecdmambavisionregionboundarytemporal}
\BIBentryALTinterwordspacing
D.~Perera, T.~Firdous, O.~Samarakoon, R.~Godaliyadda, P.~Ekanayake, and V.~Herath, ``{MambaRefine-CD}: {MambaVision} with region-boundary temporal refinement,'' 2026. [Online]. Available: \url{https://arxiv.org/abs/2607.04403}
\BIBentrySTDinterwordspacing

\bibitem{changerwkv}
Z.~Yang, G.~Pei, T.~Chen, X.~Yuan, H.~Zhang, X.~Shu, and Y.~Yao, ``Beyond quadratic: Linear-time change detection with {RWKV},'' \emph{arXiv preprint arXiv:2603.19606}, 2026.

\bibitem{wasalathilaka2026controlledbenchmarkvisualstatespace}
\BIBentryALTinterwordspacing
N.~Wasalathilaka, D.~Perera, O.~Samarakoon, B.~Wijenayake, R.~Godaliyadda, V.~Herath, and P.~Ekanayake, ``A controlled benchmark of visual state-space backbones with domain-shift and boundary analysis for remote-sensing segmentation,'' 2026. [Online]. Available: \url{https://arxiv.org/abs/2604.18721}
\BIBentrySTDinterwordspacing

\bibitem{sun2020ttt}
\BIBentryALTinterwordspacing
Y.~Sun, X.~Wang, Z.~Liu, J.~Miller, A.~A. Efros, and M.~Hardt, ``Test-time training with self-supervision for generalization under distribution shifts,'' 2019, {ICML} 2020. [Online]. Available: \url{https://arxiv.org/abs/1909.13231}
\BIBentrySTDinterwordspacing

\bibitem{gandelsman2022tttmae}
\BIBentryALTinterwordspacing
Y.~Gandelsman, Y.~Sun, X.~Chen, and A.~A. Efros, ``Test-time training with masked autoencoders,'' 2022. [Online]. Available: \url{https://arxiv.org/abs/2209.07522}
\BIBentrySTDinterwordspacing

\bibitem{wang2021tent}
\BIBentryALTinterwordspacing
D.~Wang, E.~Shelhamer, S.~Liu, B.~Olshausen, and T.~Darrell, ``Tent: Fully test-time adaptation by entropy minimization,'' 2020, {ICLR} 2021. [Online]. Available: \url{https://arxiv.org/abs/2006.10726}
\BIBentrySTDinterwordspacing

\bibitem{liang2024ttasurvey}
\BIBentryALTinterwordspacing
J.~Liang, R.~He, and T.~Tan, ``A comprehensive survey on test-time adaptation under distribution shifts,'' 2023, {International Journal of Computer Vision}, 2024. [Online]. Available: \url{https://arxiv.org/abs/2303.15361}
\BIBentrySTDinterwordspacing

\bibitem{ttt}
Y.~Sun, X.~Li, K.~Dalal, J.~Xu, A.~Vikram, G.~Zhang, Y.~Dubois, X.~Chen, X.~Wang, S.~Koyejo, T.~Hashimoto, and C.~Guestrin, ``Learning to (learn at test time): {RNNs} with expressive hidden states,'' \emph{arXiv preprint arXiv:2407.04620}, 2024.

\bibitem{schlag2021fastweights}
\BIBentryALTinterwordspacing
I.~Schlag, K.~Irie, and J.~Schmidhuber, ``Linear transformers are secretly fast weight programmers,'' 2021. [Online]. Available: \url{https://arxiv.org/abs/2102.11174}
\BIBentrySTDinterwordspacing

\bibitem{zhang2025testtimetrainingright}
\BIBentryALTinterwordspacing
T.~Zhang, S.~Bi, Y.~Hong, K.~Zhang, F.~Luan, S.~Yang, K.~Sunkavalli, W.~T. Freeman, and H.~Tan, ``Test-time training done right,'' 2025. [Online]. Available: \url{https://arxiv.org/abs/2505.23884}
\BIBentrySTDinterwordspacing

\bibitem{behrouz2024titans}
\BIBentryALTinterwordspacing
A.~Behrouz, P.~Zhong, and V.~Mirrokni, ``Titans: Learning to memorize at test time,'' 2024. [Online]. Available: \url{https://arxiv.org/abs/2501.00663}
\BIBentrySTDinterwordspacing

\bibitem{changetitans}
Z.~Yang, G.~Pei, Y.~Yao, T.~Zhou, L.~Ding, and F.~Shen, ``Towards remote sensing change detection with neural memory,'' \emph{arXiv preprint arXiv:2602.10491}, 2026.

\bibitem{vit3}
D.~Han, Y.~Li, T.~Li, Z.~Cao, Z.~Wang, J.~Song, Y.~Cheng, B.~Zheng, and G.~Huang, ``{ViT}$^3$: Unlocking test-time training in vision,'' \emph{arXiv preprint arXiv:2512.01643}, 2025, version 2, April 2026.

\end{thebibliography}
\end{document}